\documentclass[letterpaper]{article} 
\usepackage{aaai25}  
\usepackage{times}  
\usepackage{helvet}  
\usepackage{courier}  
\usepackage[hyphens]{url}  
\usepackage{graphicx} 
\usepackage{natbib}  
\usepackage{caption} 
\usepackage{algorithm}
\usepackage{algorithmic}
\usepackage{multirow}
\usepackage{amssymb}
\usepackage{newfloat}
\usepackage{listings}
\DeclareCaptionStyle{ruled}{labelfont=normalfont,labelsep=colon,strut=off} 
\floatstyle{ruled}
\newfloat{listing}{tb}{lst}{}
\floatname{listing}{Listing}
\title{Learning Cross-Domain Representations for Transferable Drug Perturbations on Single-Cell Transcriptional Responses}
\author{
    Hui Liu\thanks{Correspondence author.},
    Shikai Jin
}
\affiliations{
    College of Computer and Information Engineering, Nanjing Tech University, Nanjing, 211816, China\\

    hliu@njtech.edu.cn
}

\usepackage{bibentry}

\begin{document}

\maketitle

\begin{abstract}
Phenotypic drug discovery has attracted widespread attention because of its potential to identify bioactive molecules. Transcriptomic profiling provides a comprehensive reflection of phenotypic changes in cellular responses to external perturbations. In this paper, we propose XTransferCDR, a novel generative framework designed for feature decoupling and transferable representation learning across domains. Given a pair of perturbed expression profiles, our approach decouples the perturbation representations from basal states through domain separation encoders and then cross-transfers them in the latent space. The transferred representations are then used to
reconstruct the corresponding perturbed expression profiles via a shared decoder. This cross-transfer constraint effectively promotes the learning of transferable drug perturbation representations. We conducted extensive evaluations of our model on multiple datasets, including single-cell transcriptional responses to drugs and single- and combinatorial genetic perturbations. The experimental results show that XTransferCDR achieved better performance than current state-of-the-art methods, showcasing its potential to advance phenotypic drug discovery.
\end{abstract}

%
\begin{links}
    \link{Code}{https://github.com/hliulab/XTransferCDR}
\end{links}

\section{Introduction}

In recent years, target-based drug discovery has made significant progress ~\cite{yofe2020single,przybysla2022new}. However, the efficacy of drug molecules on their targets is profoundly influenced by the complex cellular context, and drugs designed to bind their targets do not always elicit the desired changes at the level of cellular phenotype~\cite{lim2022emerging}, partially due to off-target effects. Consequently, the success rate of target-based drug discovery remains relatively low ~\cite{sousaLuis2022drug} . These challenges have prompted researchers to increasingly focus on phenotype-based drug discovery ~\cite{vincent2022targeted}. 

Transcriptomic profiles are widely used as molecular phenotype data to elucidate cellular responses to various perturbations ~\cite{Replogle2022-ci}, such as drugs and genetic knockout/activation. Large-scale experiments have been conducted to measure a wealth of drug-induced transcriptional responses in vitro ~\cite{yang2012genomics}. For example, the L1000 platform ~\cite{Subramanian2017-oo} offers a cost-effective, high-throughput method for generating expression profiles in response to drug perturbations ~\cite{yofe2020single}. In particular, single-cell RNA sequencing (scRNA-seq) is valuable for detecting subtle transcriptional changes within individual cells, allowing for the identification of cell subpopulations that are resistant to specific drugs and facilitating phenotypic analysis at single-cell resolution~\cite{haque2017practical,levitin2018single,ding2020single}. For instance, sci-Plex~\cite{Srivatsan2020}, which integrates nuclear hashing and single-cell RNA sequencing into a single workflow for multiplex transcriptomics, has been used to quantify global transcriptional responses to hundreds of independent perturbations at the single-cell level. Despite the power of these high-throughput screening techniques, their capacity remains limited when compared to the space of all potential cell types and perturbation combinations. As such, there is an urgent need to develop computational models capable of predicting phenotype changes induced by various perturbations.

Several methods have been proposed for predicting cellular responses to drug treatments, including deep variational autoencoders ~\cite{jia2021deep}, kernelized Bayesian matrix factorization ~\cite{madhukar2019bayesian}, matrix factorization with similarity regularization ~\cite{gao2021collaborative}, and convolutional neural networks ~\cite{zhang2022cnn}. These methods employ various techniques, such as imputing drug responses through low embeddings of multiple genes ~\cite{roohani2022gears}, incorporating prior knowledge of pathway-drug associations ~\cite{chawla2022gene}, leveraging mutational signatures
~\cite{robichaux2021structure,aissa2021single} and expression profiles for prediction ~\cite{sharifi2021out,he2022context}. Despite these advancements, most machine learning models struggle to handle high-dimensional scRNA-seq data, resulting in suboptimal performance in predicting single-cell responses. To address these limitations, deep learning techniques have been increasingly applied in the analysis of scRNA-seq data in recent years. CPA ~\cite{lotfollahi2023predicting} proposes a deep encoder-decoder framework to generate interpretable drug and cell type embeddings, facilitating the cellular response prediction of novel drug dosages and drug combinations. ChemCPA ~\cite{hetzel2022predicting} extends CPA by introducing a drug molecular encoder to predict single-cell transcriptional response to unseen drugs. GEARS ~\cite{roohani2022gears} leverages a graph neural network and a knowledge graph of gene-gene relationships to model multi-gene perturbations on transcriptional responses. PerturbNet ~\cite{yu2022perturbnet} uses a deep generative model to learn a continuous mapping between the space of possible perturbations and the space of possible cell states. However, these previous studies have not focused on learning transferable perturbation representation, which results in poor performance when applied to novel drugs or cellular contexts.

To develop computational methods for predicting transcriptional responses to novel perturbations in new cellular contexts, it is essential to effectively disentangle perturbation-specific representations and basal cellular states from expression profiles induced by external stimuli. Inspired by style transfer ~\cite{lee2021dranet} and disentangled representation learning~\cite{bousmalis2016domain}, we proposed XTransferCDR, a cross-domain transfer learning framework to predict cellular response at single-cell level. To create an interpretable model, we hypothesize that in the latent pharmacotranscriptomic space, the effects of perturbations on cellular states adhere to a linear additivity rule~\cite{lotfollahi2023predicting,hetzel2022predicting}. Therefore, we introduced a cross-transfer constraint to ensure that the extracted perturbation representations can reconstruct expected cellular response when transferred across different cellular contexts. The integration of cross-domain transfer and linear modeling facilitate the learning of interpretable and transferable perturbation representations, enabling the prediction of drug responses in novel cellular contexts. We conducted extensive evaluations of our model on multiple datasets, including single-cell transcriptional responses to drugs, as well as single and combinatorial genetic perturbations. Experimental results show that our model outperforms existing state-of-the-art methods.

We believe this work has at least the following three contributions:

\begin{itemize}
\item To our knowledge, this is the first time that cross-domain disentanglement representation learning has been proposed to model perturbation-induced cellular responses at single-cell level.
\item Cross-transfer constraints empower our model to effectively learn transferable perturbation representations, enhancing the model's generalizability in predicting cellular response to novel drugs and genetic perturbations.
\item We evaluated the proposed model on multiple single-cell transcriptional response to drug and genetic perturbations, and verified that our model achieved better performance than current state-of-the-art methods.
\end{itemize}

\section{Related Works}
\subsection{Cross-Domain Style Transfer}
Cross-domain transfer is a concept in the field of computer vision, particularly in image style transfer~\cite{zhu2017unpaired,tran2019domain}, which involves separating an image representation in the latent space into two distinct factors: content and style. Several methods have been proposed to utilize convolutional neural networks, bilinear models, and adversarial learning to effectively disentangle the content and style of images. Our method is primarily inspired by cross-domain disentanglement network ~\cite{gonzalez2018image} and DRANet ~\cite{lee2021dranet}, which employ a cross-domain autoencoder framework to disentangle images into content (scene structure) and style (artistic appearance), realizing bidirectional image-to-image translation. Pioneering work informs our approach to learning transferable perturbation representations, which facilitate the prediction of transcriptional responses.

\subsection{Linear Modeling in Latent Space}
The linear additive model in the latent space is widely used in deep learning for interpretability, such as latent additive neural models~\cite{nguyen2023flan} and latent linear additivity models~\cite{lotfollahi2023predicting,hetzel2022predicting,huang2024predicting}. CPA \cite{lotfollahi2023predicting} and cycleCDR \cite{huang2024predicting} are most related to our work, as they combine the interpretability of linear models with the power of deep learning to model single-cell transcriptional responses. Linear modeling in latent space yields easy-to-interpret embeddings that represent drug-induced perturbations within the cellular context, thereby promoting the discovery of drugs with similar therapeutic effect and synergistic drug combinations.

\begin{figure*}[t]
\centering
\includegraphics[width=0.9\textwidth]{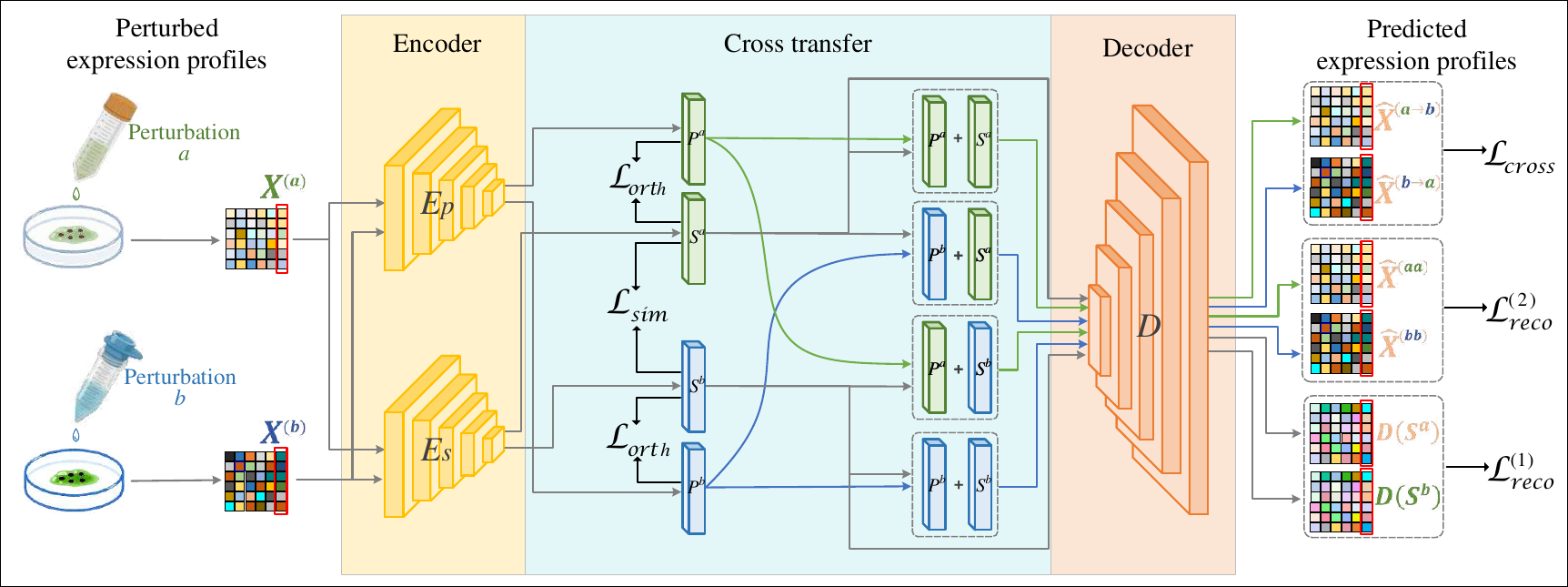} 
\caption{Illustrative diagram of the proposed XTransferCDR framework. Two encoders $E_s$ and $E_p$ extract basal state and perturbation embeddings from perturbed expression profiles, and cross transfer the perturbation embeddings in the latent space, and a shared decoder $D$ map them back to reconstruct corresponding expression profiles.}
\label{fig:framework}
\end{figure*}

\section{XTransferCDR}
\subsection{Framework Overview}
Given the expression profiles perturbed by drugs or genetic manipulations, our goal is to learn perturbation representations transferable to novel cellular contexts. For this purpose, we introduce feature disentangling learning to decompose the expression profiles induced by two distinct perturbations into their respective perturbation representations and cellular basal states. Meanwhile, we postulate that the effect of external perturbations on cellular states follows linear additivity in the latent space, as adopted by previous studies\cite{hetzel2022predicting,huang2024predicting}. As such, we cross-transfer the perturbation representations in the latent space to reconstruct the corresponding perturbed expression profile through a decoder.

Figure~\ref{fig:framework} illustrates the architecture of our proposed XTransferCDR model, which primarily consists of four components: an encoder $E_s$ used to extract cellular basal state, and an encoder $E_p$ to extract perturbation embeddings from the perturbed gene expression profiles, a cross-transfer module to swap these two perturbation embeddings and recombine with respective basal state, and a decoder $D$ to map the embeddings in the latent space back to corresponding expression profiles. In principle, we borrow the idea from disentangled representation learning and style transfer in conceptualizing the XTransferCDR framework.

\subsection{Representation Disentanglement}
Suppose that an unperturbed expression profile is denoted $X$, and the expression profiles induced by perturbations $a$ and $b$ are represented as $X^{(a)}$ and $X^{(b)}$, respectively. The objective of the encoder $E_s$ is to extract the cellular basal state that can be mapped back by a decoder to reconstruct the unperturbed expression profile $X$. We assume that the basal states extracted from $X^{(a)}$ and $X^{(b)}$ are denoted by $S^{(a)}=E_s(X^{(a)})$ and $S^{(b)}=E_s(X^{(b)})$ in the latent space. Meanwhile, the encoder $E_p$ is designed to extract perturbation representations from perturbed expression profiles. Formally, the encoder $E_p$ extracts the respective perturbation embeddings from $X^{(a)}$ and $X^{(b)}$, denoted by $P^{(a)}=E_p(X^{(a)})$ and $P^{(b)}=E_p(X^{(b)})$ in the latent spaces. For feature disentanglement, we enforce an orthogonality constraint between the basal state and perturbation representations extracted from the same perturbed expression profiles. For example, $P^{(a)}$ and $S^{(a)}$, as well as $P^{(b)}$ and $S^{(b)}$, are orthogonal to each other so that they do not contain redundant information. Formally, the orthogonal loss is defined as follows:
\begin{equation}
    \mathcal{L}_{orth}=\frac{1}{N}\sum_i \left\|P^{(a)}_i\cdot S^{(a)}_i\right\|_F^2+ \left\|P^{(b)}_i\cdot S^{(b)}_i\right\|_F^2
\end{equation}
in which $\left\|.\right\|_F^2$ was the squared Frobenius norm, $i$ is the index of paired expression profiles in a mini-batch. The orthogonality constraints enforce that the basal encoder and perturbation encoder actually function as a domain separation network ~\cite{bousmalis2016domain} to extract disentangled representations. Specifically, once the basal state is aligned, the perturbations can be cross-transferred to recover the corresponding perturbed expression profiles through a decoder.

Moreover, if two perturbations $a$ and $b$ are applied to the same cellular context, the basal states $S^{(a)}$ and $S^{(b)}$ extracted from the respective perturbed expression profiles should be aligned to each other as much as possible. Therefore, we try to minimize the similarity between them, and the Kullback-Leibler divergence is used to define the similarity loss $\mathcal{L}_{sim}$ as follows:
\begin{equation}
 \mathcal{L}_{sim}=\sum_i E_{s}(X^{(a)}_i)\log\frac{E_{s}(X^{(a)}_i)}{E_{s}(X^{(b)}_i)}
\end{equation}

\subsection{Expression Profile Reconstruction}
We expect that the basal states $S^a$ and $S^b$ extracted by $E_s$ can reconstruct the unperturbed expression profile $X$ by the decoder $D$. For this purpose, we define the mean square error between the actual unperturbed expression profile and the inferred ones as the reconstruction loss, which should be minimized during model training. Formally, we define the reconstruction loss as follows:
\begin{equation}
    \mathcal{L}_{reco}^{(1)}=\frac{1}{N}\sum_{i=1}^{N}\left(D(S^{(a)}_i)-X_i\right)^{2}+\left(D(S_i^{(b)})-X_i\right)^{2}
\end{equation}
Because the basal state is aligned to the unperturbed expression profiles, the effect of perturbation leading to phenotypic change is completely encoded by the perturbation embeddings.

Moreover, we hypothesize that the effect of perturbations on the cellular state follows a linear additive rule in the latent space, so that we can recover the perturbed expression profile by the same decoder. Specifically, since the expression profiles $X^{(a)}$ are orthogonally decoupled into the basal state $S^{(a)}$ and perturbation feature $P^{(a)}$, the sum of $S^{(a)}$ and $P^{(a)}$ contains all the information needed to reconstruct the perturbed expression profiles. Therefore, the decoder $D$ is assumed to take ($P^{(a)} + S^{(a)}$) as input and output the expression profile $X^{(a)}$. The same operation is applied to $X^{(b)}$. Formally, denote by the inferred expression profiles $\hat{X}^{(aa)}=D(S^{(a)}+P^{(a)})$ and $\hat{X}^{(bb)}=D(S^{(b)}+P^{(b)})$, we define the mean square error between the actual and inferred perturbed expression profiles as another reconstruction loss:
\begin{equation}
    \mathcal{L}_{reco}^{(2)}=\frac{1}{N}\sum_i^N \left(\hat{X}^{(aa)}_i-X^{(a)}_i\right)^{2}+\left(\hat{X}^{(bb)}_i-X^{(b)}_i\right)^{2}
\end{equation}
We think the reconstruction losses offer at least some significance. First, the basal state encoder is enforced to capture the ground-truth information related to the unperturbed cellular state. Second, the perturbation encoder focuses on extracting the complete information relevant to perturbation. Finally, the decoder is endeavored to  reconstruct corresponding expression profiles given the established representation in the latent space.

\subsection{Cross Transfer}
To learn transferable perturbation representations, we swap the representations $P^{(a)}$ and $P^{(b)}$ extracted from the expression profiles $X^{(a)}$ and $X^{(b)}$ induced by two different perturbations in the same cellular context. Under the assumption of linear additivity rule, the cross-transferred perturbations were respectively added to the basal states $S^{(a)}$ and $S^{(b)}$ to simulate the effect of perturbations on the cellular state, namely ($P^{(a)}+S^{(b)}$) and ($P^{(b)}+S^{(a)}$). Next, the shared decoder $D$ is assumed to recover the corresponding perturbed expression profiles. Denote by $\hat{X}^{(a\rightarrow b)}=D(S^{(b)}+P^{(a)})$ and $\hat{X}^{(b\rightarrow a)}=D(S^{(a)}+P^{(b)})$, we define the mean square error loss function as follows:
\begin{equation}
    \mathcal{L}_{cross}=\frac{1}{N}\sum_i^N \left(\hat{X}^{(a\rightarrow b)}_i-X^{(a)}_i\right)^{2}+\left(\hat{X}^{(b\rightarrow a)}_i-X^{(b)}_i\right)^{2}
\end{equation}

Overall, the orthogonal constraint enforces two encoders $E_s$ and $E_p$ function collaboratively to realize feature disentanglement, while the cross-transfer constraint enforces them to learn the transferable perturbation representations. 

\subsection{Full Objective}
The full objective function is defined as below:
\begin{equation}
\mathcal{L}=\mathcal{L}_{sim}+\mathcal{L}_{orth}+\mathcal{L}_{reco}^{(1)}+\mathcal{L}_{reco}^{(2)}+\mathcal{L}_{cross}
\end{equation}
The total loss function is minimized to learn the parameters of the encoders and decoder.

In our practice, the encoder/decoder are realized using fully-connected feed-forward networks with rectified linear unit (ReLU) activation function. They consist of four feed-forward layers with sizes of 1024, 512, 256, and 128, respectively. Each feed-forward layer is followed by a batch normalization layer, and a dropout layer with the dropout probability set to 0.2. The learning rate is set to 2e-4, and the bottleneck dimension between the encoder and decoder is set to 128. The model was trained for 60 epochs, and all experiments were conducted on a CentOS Linux 8.2.2004 (Core) system, equipped with a GeForce RTX 4090 GPU and 128GB memory. During the model training and cross-validation stage, these loss terms were appropriately weighted.

\begin{table*}[t]
\centering
\begin{tabular}{|c|c|c|c|c|}
\hline
Model  & mean $R^2$ (All genes) & mean $R^2$ (DEGs)  & median $R^2$ (All genes) & median $R^2$ (DEGs) \\ \hline
Baseline        & 0.50    & 0.29   & 0.49  & 0.12  \\ \hline
chemCPA         & 0.51    & 0.32   & 0.47  & 0.24  \\ \hline
\begin{tabular}[c]{@{}c@{}}chemCPA+pretraining\end{tabular}
                & 0.68    & 0.54   & 0.75  & 0.64  \\ \hline
cycleCDR        & 0.72    & 0.55   & 0.80  & 0.69  \\ \hline
XTransferCDR  & \textbf{0.81}  & \textbf{0.62} & \textbf{0.90}  & \textbf{0.72}  \\ \hline
\end{tabular}
\caption{Performance evaluation on the sci-Plex3 single-cell transcriptional response dataset.}
\label{tab:sciplex3}
\end{table*}

\begin{table*}[t]
\centering
\begin{tabular}{|c|c|c|c|c|c|c|}
\hline
Model & $R^2$ (All genes) & $R^2$ (DEGs) & EV (All genes) & EV (DEGs) & PCC (All genes) & PCC (DEGs) \\ \hline
Baseline & 0.432 & 0.166 & 0.435 & 0.168 & 0.348 & 0.189 \\ \hline
cycleCDR & 0.620 & 0.629 & 0.637 & 0.750 & 0.827 & 0.894 \\ \hline
XTransferCDR & \textbf{0.861} & \textbf{0.806} & \textbf{0.865} & \textbf{0.822} & \textbf{0.932} & \textbf{0.911} \\ \hline
\end{tabular}
\caption{Performance Comparison on the sci-Plex4 single-cell transcriptional response to drug perturbations.}
\label{tab:sciplex4}
\end{table*}

\section{Experiments}
\subsection{Experimental Setup}
During the training stage, our model requires paired samples as input, namely, pairs of expression profiles induced by two distinct perturbations. To achieve this, given the single-cell expression profiles of certain cell lines subjected to a large number of perturbations, they were divided into two groups with approximately equal sizes at the drug level. Next, we randomly selected one case from each group to form a paired sample till all cases were paired. The drug-level data partition ensures that each paired sample corresponds to two different drugs, thereby avoiding symmetric repetition. 

For objective performance evaluation, each established dataset was randomly divided into training, validation, and test sets. Importantly, all data points associated with a specific drug were exclusively allocated to either the training or test set. In the testing stage, the perturbed expression profile by a perturbation ``unseen" in training stage was fed into the encoder to extract its perturbation representation, which is then combined with the basal state derived  from corresponding unperturbed expression profiles. The resulting embedding is used to infer the expression profile induced by this ``unseen" perturbation. The predicted expression profiles were subsequently compared to the actual ones to assess the model performance. This testing strategy was intentionally designed to evaluate the model's ability to generalize to novel drugs when predicting cellular transcriptional responses.

\begin{figure}[t]
\centering
\includegraphics[width=0.9\columnwidth]{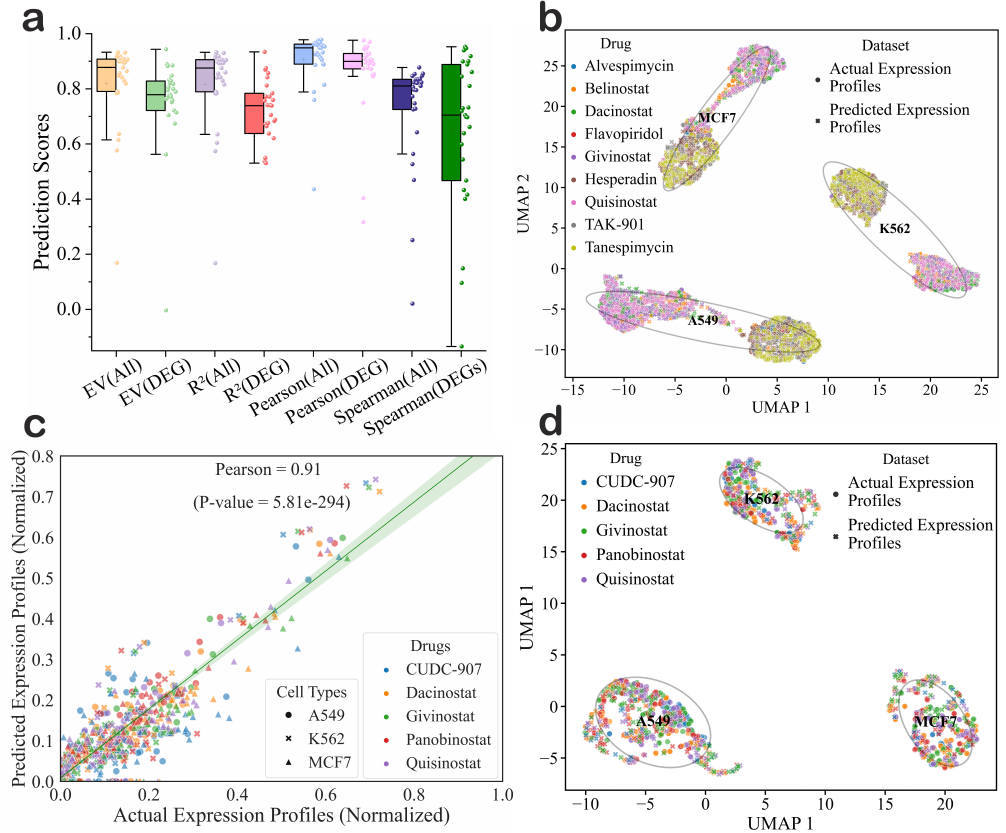} 
\caption{Performance evaluation on sci-plex single-cell transcriptional response to drug perturbations. (a) Performance metrics acheived by XTransferCDR on sci-plex3 dataset. (b) UMAP visualization of predicted and actual expression profiles induced by 9 drugs with most significant effects on sci-plex3 dataset. (c) Scatter plot between predicted and actual expression profiles on sci-plex4 dataset. (d) UMAP visualization of predicted and actual expression profiles induced by 5 drugs with most significant effects on sci-plex4 dataset.}
\label{fig:plex3}
\end{figure}

\subsection{Evaluation Metrics}
We calculated three performance metrics: coefficient of determination (R$^2$), Pearson correlation coefficient (PCC), and explained variance (EV) using the predicted and actual expression profiles. Given that most genes exhibit minimal change in response to external perturbations, the performance metrics computed across genome-wide expression profiles may not accurately reflect the model's true predictive capability. Instead, the differentially expressed genes (DEGs) could more faithfully capture the actual impact of perturbations on cellular states. Therefore, we also reported the performance metrics computed across the top 50 DEGs. Note that the differential expression analysis was conducted by comparing the perturbed expression levels to the unperturbed ones, with the threshold for differential expression defined as a logarithmic fold change greater than 1 ($\log_2|fc|\geq 1$). Additionally, to evaluate our model's performance, we compared it against previously published methods, including chemCPA ~\cite{hetzel2022predicting}, GEARS ~\cite{roohani2022gears} and cycleCDR~\cite{huang2024predicting}. We also introduced a baseline model to serve as a benchmark. The baseline model calculates performance metrics directly using the unperturbed and actual perturbed expression profiles.

\subsection{Evaluation on Single-Cell Drug Response}
We initially evaluated the model's performance on the single-cell chemical response dataset from the sci-Plex project~\cite{Srivatsan2020}. The sci-Plex3 dataset contains the single-cell transcriptional responses of three human cancer cell lines (MCF7, K562, and A549) exposed to 188 different drugs. To capture pronounced drug effects, we adopted the expression profiles induced by a 10µM drug dosage. Among the drugs, nine drugs (Dacinostat, Givinostat, Belinostat, Hesperadin, Quisinostat, Alvespimycin, Tanespimycin, TAK-901, and Flavopiridol) were reported in the original sci-Plex publication as the most effective in inducing significant change of expression levels. These nine drugs involve in three distinct mechanisms of action, including epigenetic regulation, tyrosine kinase signaling, and cell cycle regulation. For model evaluation, the expression profiles induced by these nine drugs were held out as the test set (n=3,071), while the remaining data were used to create the paired samples for training ($n$=101,190) and validation set ($n$=8,499) with a 4:1 ratio. 

\begin{table*}[t]
\centering
\begin{tabular}{|c|c|c|c|c|c|}
\hline
Cell line & Method & $R^2$ (All genes) & $R^2$ (DEGs) & EV (All genes) & EV (DEGs) \\ \hline
\multirow{4}{*}{K562} & Baseline & 0.84 & 0.10 & 0.84 & 0.19 \\ \cline{2-6}
& GEARS & 0.86 & 0.29 & 0.86 & 0.32 \\ \cline{2-6}
& cycleCDR  & 0.88 & 0.37 & 0.88 & 0.40 \\ \cline{2-6}
& XTransferCDR & \textbf{0.90} & \textbf{0.68} & \textbf{0.90} & \textbf{0.72} \\ \hline 
\multirow{4}{*}{RPE-1} & Baseline & 0.72 & 0.00 & 0.72 & 0.11 \\ \cline{2-6}
& GEARS & 0.80 & 0.27 & 0.81 & 0.30 \\ \cline{2-6}
& cycleCDR  & 0.82 & 0.45 & 0.82 & 0.46 \\ \cline{2-6}
& XTransferCDR & \textbf{0.86} & \textbf{0.78} & \textbf{0.86} & \textbf{0.80} \\ \hline
\end{tabular}
\caption{Performance evaluation on single-cell response datasets induced by single-gene genetic perturbations.}
\label{tab:gears}
\end{table*}

For each cell in the test set subjected to a specific drug perturbation, we predicted its perturbed expression profile, and then calculated the performance metrics using the actual and predicted mean expression levels of all cells across all genes and top 50 DEGs, respectively. To benchmark the performance, we compared our method to baseline and cycleCDR ~\cite{huang2024predicting}, as well as chemCPA ~\cite{hetzel2022predicting} with and without pretraining on L1000 bulk data. Table \ref{tab:sciplex3} presented the mean and median $R^2$ values on the hold-out test set. It can be found that our model achieved significantly higher performance than other comparative methods. Figure~\ref{fig:plex3}(a) presented the boxplots of other performance metrics achieved by our method, computed across all genes and DEGs. Our method consistently exhibited superior performance on these metrics, with the exception of a relatively large variance in the Spearman correlation coefficient for DEGs. To facilitate visual inspection, we employed the UMAP tool to visualize the actual and predicted gene expression profiles. As illustrated in Figure~\ref{fig:plex3}(b), the predicted expression profiles closely aligned with the actual ones, further validating the accuracy of our model.

Moreover, we have noticed that the sci-Plex project has released a new dataset sci-Plex4, thereby we extended our evaluation to this dataset. Following the drug-level data partitioning strategy, the sci-Plex4 dataset was randomly divided into a training set ($n$=7,104), validation set ($n$=718), and test set ($n$=733). The test set included five distinct drugs: Dacinostat, CUDC-907, Quisinostat, Panobinostat, and Givinosta. In Table \ref{tab:sciplex4}, we presented the means of $R^2$, EV, and PCC metrics on the hold-out test set. As expected, our model consistently outperformed the baseline model across all evaluation metrics. Figure~\ref{fig:plex3}(c-d) illustrated strong correlation between the actual and predicted expression profiles, and the UMAP plot verified that the predicted expression profiles aligned closely to the actual ones.

\subsection{Evaluation on Single-Cell Genetic Perturbation}
We further evaluated our method on two single-cell datasets established by genetic perturbation assays ~\cite{Replogle2022-ci}. The datasets contain the single-cell RNA sequencing readouts of the entire transcriptome of K562 and RPE-1 cells through CRISPR-based knockout of individual genes. Following aforementioned data partition strategy, the K562 dataset was divided into a training set (n=64,249), a validation set ($n$=2,234), and a test set ($n$=2,233). The RPE-1 dataset was divided into a training set ($n$=72,200), a validation set ($n$=2,045), and a test set ($n$=2,044). 

\begin{figure}[t]
\centering
\includegraphics[width=0.9\columnwidth]{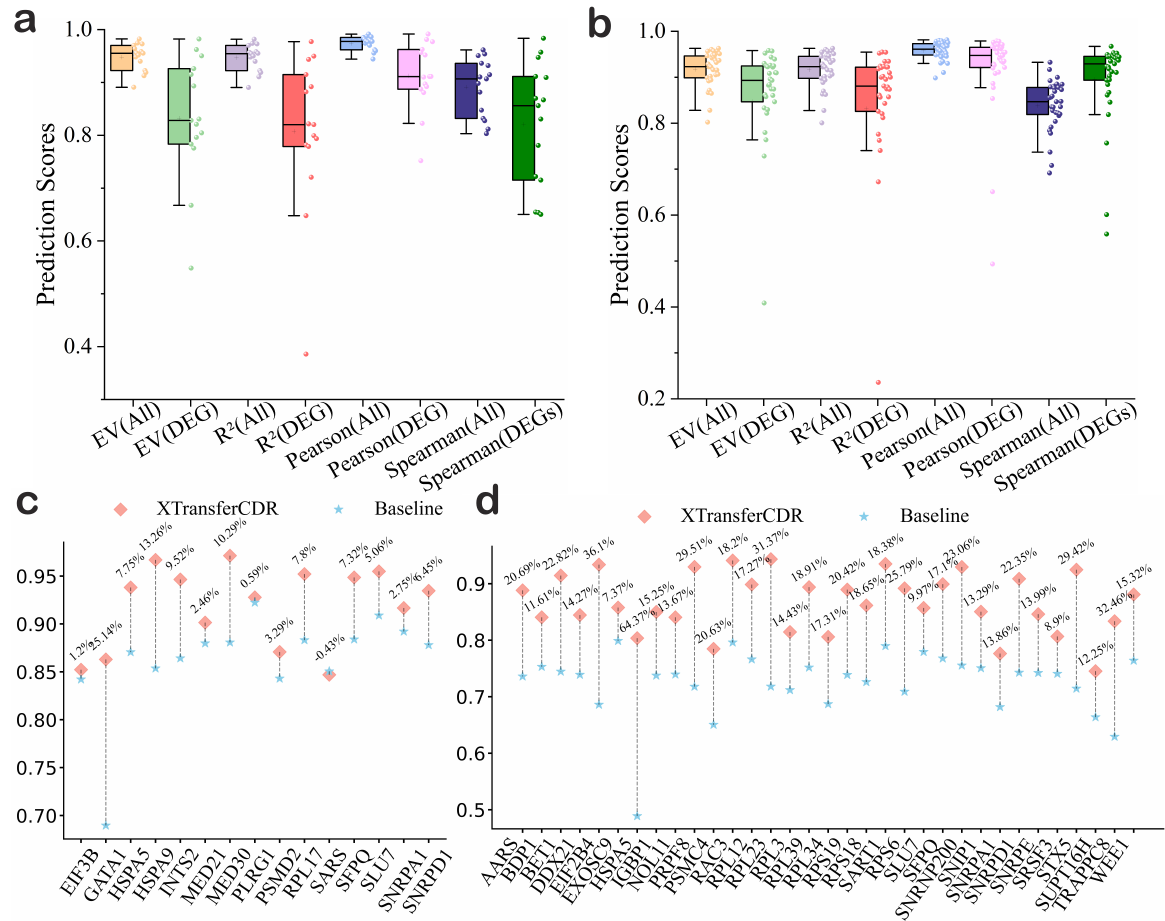} 
\caption{Performance evaluation on single-cell transcriptional response to genetic perturbations on k562 and RPE-1 cell lines. (a-b) Performance metrics achieved by XTransferCDR across all genes and differentially expressed genes (DEGs), respectively. Although the metrics computed on DEGs on k562 and RPE-1 cell lines, respectively. (c-d) Performance advantage over baseline model for the top 5\% knockout genes that induced most significant differential expression on k562 and RPE-1 cell lines, respectively.}
\label{fig:gears}
\end{figure}

In addition to the baseline model, we conducted performance comparison with GEARS ~\cite{roohani2022gears} and cycleCDR ~\cite{huang2024predicting}. Table~\ref{tab:gears} showed the mean $R^2$ and EV values on the hold-out test sets. Notably, our model outperformed all comparative methods across all performance metrics. Moreover, Figure~\ref{fig:gears}(a-b) showed other performance metrics achieved by our method on K562 and RPE-1 datasets, respectively. Although the metrics computed on DEGs exhibited relatively large variance, our method always achieved more than 0.8 mean performance measures. To explore the performance on specific genetic perturbations, we paid close attention to the top 5\% of knockout genes that induced most significant differential expression. As shown in Figures \ref{fig:gears}(c-d), our method exhibited notable performance advantage over the baseline model on these knockout genes, except for the SARS gene in RPE-1 cell line. These findings confirm the exceptional performance and generalization ability of our model in predicting transcriptional responses induced by gene perturbations. 

\begin{table*}[t]
\centering
\setlength{\tabcolsep}{4pt} 
\begin{tabular}{|c|c|c|c|c|c|c|}
\hline
Model & \multicolumn{1}{l|}{$R^2$ (All genes)} & $R^2$ (DEGs) & EV (All genes) & EV (DEGs) & PCC (All genes) & PCC (DEGs) \\ \hline
Baseline & 0.852 & 0.166 & 0.859 & 0.398 & 0.927 & 0.711 \\ \hline
XTransferCDR & \textbf{0.964} & \textbf{0.837} & \textbf{0.965} & \textbf{0.877} & \textbf{0.983} & \textbf{0.877} \\ \hline
\end{tabular}
\caption{Performance evaluation on single-cell response dataset induced by combinatorial genetic perturbations.}
\label{tab:dual-gene}
\end{table*}

\begin{table*}[t]
\centering
\setlength{\tabcolsep}{3pt} 
\begin{tabular}{|c|c|c|c|c|c|c|c|c|}
\hline
$\mathcal{L}_{cross}$ & $\mathcal{L}_{orth}$ & $\mathcal{L}_{sim}$ & $R^2$ (All genes) & $R^2$ (DEGs) & EV (All genes) & EV (DEGs) & PCC (All genes) & PCC (DEGs) \\ \hline
  & \checkmark & \checkmark & 0.660 & 0.183 & 0.663 & 0.270 & 0.814 & 0.575 \\ \hline
\checkmark &  & \checkmark & 0.724 & 0.521 & 0.729 & 0.677 & 0.850 & 0.823 \\ \hline
\checkmark & \checkmark &  & 0.752 & 0.521 & 0.755 & 0.615 & 0.857 & 0.782 \\ \hline
\checkmark & \checkmark & \checkmark & \textbf{0.817} & \textbf{0.626} & \textbf{0.819} & \textbf{0.717} & \textbf{0.905} & \textbf{0.861} \\ \hline
\end{tabular}
\caption{Performance evaluation of ablated models on sci-plex3 dataset.}
\label{tab:ablation}
\end{table*}

\subsection{Evaluation on Combinatorial Genetic Perturbations}
To further validate the effectiveness of learned transferable perturbations, we carried out systematic evaluation on another dataset that was generated through CRISPR-based knockout (deactivation) of multiple genes, aimed at observing the consequent alterations in single-cell phenotypes~\cite{norman2019exploring}. This dataset comprises the single-cell expression profiles of A549 cells subjected to  131 dual-gene knockouts and 85 distinct single-gene knockouts. Among these, we selected 10 representative dual-gene perturbations, such as (FOSB+CEBPB) and (ZBTB10+DLX2), as the test set. Notably, these dual-gene perturbations were also chosen for testing in the GEARS study \cite{roohani2022gears}, thus justifying a robust and objective performance comparison.

We trained our model on the expression profiles induced by single-gene knockouts. Next, the trained encoders were used to extract the representations of single-gene perturbations (e.g., $P^{(a)}$ and $P^{(b)}$). Following the assumption of linear additivity in the latent space, these perturbation representations were linearly combined with the basal state $S$ of the test cells to simulate the effects of dual-gene perturbation (e.g., $S+P^{(a)}+P^{(b)}$). Finally, the decoder was used to predict the expression profiles resulting from the dual-gene perturbation. For instance, given the expression profiles induced independently by single-gene knockouts (e.g., FOSB and CEBPB) in A549 cell line, we can predict the expression profiles induced by the dual-gene perturbation (FOSB+CEBPB) in the same cell line.

\begin{figure}[t]
\centering
\includegraphics[width=0.9\columnwidth]{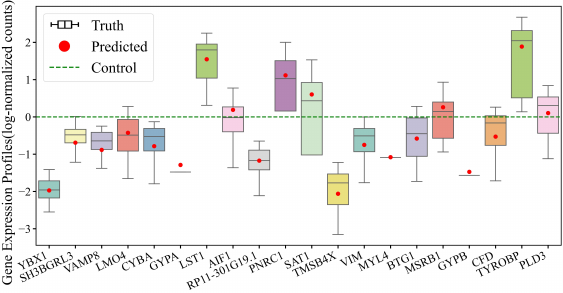} 
\caption{Predicted single-cell expression levels of top 20 differentially expressed genes induced by dual-gene perturbations (FOSB+CEBPB) on A549 cell line.}
\label{fig:dual-gene}
\end{figure}

Table~\ref{tab:dual-gene} presented the performance metrics of XTransferCDR compared to baseline model, demonstrating that our method exhibited superior predictive power for combinatorial genetic perturbations. Especially, our method exhibited great advantage over baseline model in predicting perturbed profiles of DEGs, i.e., the $R^2$, EV and PCC metrics computed across DEGs are raised from 0.166, 0.398 and 0.711 to 0.837, 0.877 and 0.877, respectively. Furthermore, we conducted an in-depth analysis to evaluate our model's performance in predicting the transcriptional response of individual genes. For example, following the dual-gene (FOSB+CEBPB) perturbation, XTransferCDR accurately predicted the expression level changes of top 20 differentially expressed genes relative to the control state, as illustrated in Figure~\ref{fig:dual-gene}. These findings further validated the reliability of our method in dissecting complex perturbations in single-cell cellular context.

\subsection{Model Ablations}
To validate the contributions of key components to the model's performance, we conducted a few ablation experiments focusing on the cross-transfer loss $\mathcal{L}_{cross}$, orthogonality constraint $\mathcal{L}_{orth}$, and basal state similarity $\mathcal{L}_{sim}$. For simplicity, we did not include the ablation of the reconstruction losses, since they were implicitly ensured by autoencoder. The ablation experiments were performed on the sci-Plex3 dataset, applying the aforementioned dataset partition strategy for objective performance evaluation.

As shown in Table~\ref{tab:ablation}, the exclusion of cross-transfer loss resulted in notable decline in performance. Specifically, the $R^2$, EV, and PCC metrics computed across all genes dropped to 0.660, 0.663, and 0.814, respectively. Conversely, the inclusion of cross-transfer loss raised the corresponding metrics to 0.817, 0.819, and 0.905, respectively. In particular, the most significant improvement was observed in the performance metrics computed across differentially expressed genes (DEGs), with $R^2$, EV, and PCC increased by 44\%, 43\%, and 28\%, respectively. This substantial enhancement verified the key contribution of cross transfer in boosting model performance.

We also examined the impact of orthogonality constraint on model optimization. This constraint enforced the disentanglement of perturbations  from basal state, thereby facilitating cross transfer between perturbation representations. As expected, its inclusion yielded significant performance improvement. As shown in Table~\ref{tab:ablation}, the $R^2$ computed across all genes and DEGs increased from 0.72 and 0.52 to 0.81 and 0.62, respectively. Furthermore, our ablation experiments highlighted the effectiveness of basal state similarity constraint, albeit its impact was not equally pronounced compared to the cross-transfer loss. When all loss terms were combined, our model achieved optimal performance. These ablation experiments demonstrated that various components functioned complementarily and collectively contribute to the overall model performance.

\section{Conclusion}
In this paper, we proposed a novel cross-domain representation learning framework to predict cellular response to external perturbations at single-cell level. We introduced a cross-transfer constraint in the latent space to ensure that the learned perturbation representations can accurately reconstruct the expected cellular responses when swapped to novel cellular contexts. By combining cross-domain transfer with linear modeling, our approach can learn interpretable and transferable perturbation representations. We rigorously evaluated our model across multiple datasets, including single-cell transcriptional responses to drugs, and both single and combinatorial genetic perturbations. The experimental results demonstrate that our model surpasses the performance of current state-of-the-art methods.

\setcounter{table}{0}
\renewcommand{\thetable}{S\arabic{table}}
\setcounter{figure}{0}
\renewcommand{\thefigure}{S\arabic{figure}}

\section{Acknowledgments}
This work was supported by National Natural Science Foundation of China
(No. 62072058, No. 62372229), Natural Science Foundation of Jiangsu
Province (No. BK20231271).

\bibliography{reference}

\end{document}